%% file: main.tex
\documentclass[11pt,a4paper]{article}

\usepackage{placeins}
\usepackage{float}

\usepackage[margin=1in]{geometry}
\usepackage{graphicx}
\usepackage{booktabs}
\usepackage{amsmath,amssymb}
\usepackage{siunitx}
\usepackage{hyperref}
\usepackage{cleveref}
\usepackage{authblk}
\usepackage{caption}
\usepackage{subcaption}
\usepackage{float}
\usepackage{xcolor}
\usepackage{microtype}
\usepackage[numbers,sort&compress]{natbib}

\graphicspath{{figures/}}

\title{Interpretable Temporal Facial-Region Motion Analysis for In-the-Wild Parkinson’s Disease Video Classification}

\author[1]{Riyadh Almushrafy}
\affil[1]{Department of Computer Science and Information, College of Science, Majmaah University, Saudi Arabia}

\date{}

\begin{document}
\maketitle

\input{sections/abstract}
\input{sections/keywords}

\input{sections/01_introduction}
\input{sections/02_related_work}
\input{sections/03_materials_methods}
\input{sections/04_experimental_setup}
\input{sections/05_results}
\input{sections/06_discussion}

\input{sections/07_limitations}
\input{sections/08_conclusion}
\input{sections/data_code_availability}
\input{sections/conflict_of_interest}

\input{sections/acknowledgments}

\bibliographystyle{plainnat}
\bibliography{bib/references}

\end{document}

%% file: sections/abstract.tex
\begin{abstract}
Reduced facial expressivity is a common motor manifestation of Parkinson's disease (PD), often described as hypomimia or facial bradykinesia. Although video-based facial analysis has been studied in clinical and semi-controlled settings, less is known about which temporal facial-region descriptors are most informative in unconstrained video. This question is particularly relevant for YouTubePD, where public videos vary in pose, lighting, speaking behavior, head motion, and recording quality.

This study investigates temporal facial-region dynamics using the region-level keypoints provided with the YouTubePD benchmark. Each video is represented through geometric descriptors extracted from 14 predefined facial regions. Static geometry, normalized geometry, velocity-based descriptors, relative-velocity descriptors, and a GRU sequence baseline are compared under a consistent binary PD-related classification protocol. Seed-robustness, region-level ablation, and permutation-importance analyses are then used to examine the stability and interpretability of the strongest representation.

The best performance was obtained with normalized velocity descriptors and a Random Forest classifier, reaching a balanced accuracy of 0.826 and AUROC of 0.855 on the held-out YouTubePD test split. Across 10 random seeds, the same representation remained stable, with balanced accuracy of $0.810 \pm 0.018$ and AUROC of $0.855 \pm 0.005$. Region-level ablation and permutation importance showed that discriminative information was unevenly distributed across the YouTubePD facial regions and was driven mainly by variability and peak changes in region-level motion.

These findings suggest that normalized temporal facial-region motion provides a lightweight and interpretable representation for in-the-wild PD-related video classification on YouTubePD. The study is presented as a benchmark-level representation analysis and does not claim medication-controlled clinical severity assessment or MDS-UPDRS facial-expression scoring.
\end{abstract}

%% file: sections/keywords.tex

%% file: sections/01_introduction.tex
\section{Introduction}
\label{sec:introduction}

Parkinson's disease (PD) can affect facial motor behavior in ways that are clinically visible but difficult to quantify consistently. Hypomimia and facial bradykinesia are commonly associated with reduced spontaneous movement, diminished expression amplitude, and lower variability during deliberate or natural facial actions. Because these symptoms are often assessed by observation, video-based facial analysis has been investigated as a way to describe PD-related facial behavior more objectively~\cite{abrami2021hypomimia,bandini2017facial,jin2020facial_expression_pd,pegolo2022hypomimia,gomez2023action_units_pd,novotny2022facialbradykinesia,su2021hypomimia}.

Most work on PD facial analysis has been developed around clinical or semi-controlled recordings, where task instructions, camera placement, and clinical context are known. These settings are appropriate for clinical assessment, but they differ from public in-the-wild videos, where facial behavior is mixed with pose variation, lighting changes, speaking behavior, head movement, camera framing, and variable video quality. Such variability makes the problem more difficult, but it also reflects the conditions under which scalable video-based screening or cue analysis may eventually need to operate.

The YouTubePD benchmark provides a public setting for studying PD-related cues in unconstrained video~\cite{zhou2023youtubepd,zhou2023youtubepd_supplement}. It includes in-the-wild videos, audio, facial landmarks, region-level annotations, and expert-provided labels. Its public availability makes it valuable for reproducible benchmarking, but its clinical interpretation must be handled carefully. Information such as medication ON/OFF state, time since medication, standardized facial task condition, disease duration, and detailed clinical rating context is not generally available. For this reason, YouTubePD is better treated as a benchmark for in-the-wild PD-related video classification and facial cue analysis, rather than as a medication-controlled clinical severity dataset.

Within this setting, an important question remains: which facial-region representations are most useful for PD-related classification? Appearance-based and multimodal models can exploit rich visual and audio information, but their predictions may be difficult to interpret and may be influenced by dataset-specific factors. Static facial geometry can describe facial configuration, yet hypomimia and facial bradykinesia are inherently dynamic. They are expressed not only in how the face appears at a given frame, but also in how facial regions move, vary, and change over time~\cite{bandini2017facial,novotny2022facialbradykinesia}. This motivates a focused analysis of temporal facial-region descriptors.

This study investigates the facial-region keypoint representation provided with YouTubePD. Each video is described using geometric descriptors extracted from 14 predefined facial regions. From these descriptors, we derive raw geometric features, frame-normalized features, static summaries, velocity-based descriptors, and relative-velocity descriptors designed to reduce shared global movement. These representations are compared using logistic regression, support vector machine, Random Forest, and a GRU-based sequence baseline under a consistent binary classification protocol~\cite{breiman2001randomforests,cortes1995support,cho2014gru}.

The aim is not to propose a clinical diagnostic system or estimate medication-controlled PD severity. Instead, the study asks which temporal facial-region descriptors provide useful and interpretable PD-related cues within the public YouTubePD benchmark. This distinction is central to the design of the work: the analysis is intended to support reproducible representation-level understanding, while avoiding clinical claims that would require standardized examination protocols and richer clinical metadata.

The contributions are threefold. First, the paper develops a lightweight temporal facial-region dynamics pipeline that converts YouTubePD region-level keypoints into normalized geometric and velocity-based descriptors. Second, it provides a systematic comparison of raw, normalized, static, velocity, relative-velocity, and GRU-based representations for in-the-wild PD-related binary classification. Third, it examines interpretability and stability through seed-robustness, region-level ablation, and permutation-importance analyses, identifying which region, descriptor, and temporal-statistic groups contribute most strongly to the classification signal.

Overall, this work positions normalized facial-region velocity as an interpretable and reproducible representation for YouTubePD-based PD video classification. The results are intended to clarify the role of temporal facial-region dynamics in an in-the-wild benchmark, rather than to replace controlled clinical assessment.

%% file: sections/02_related_work.tex
\section{Related Work}
\label{sec:related_work}

\subsection{Facial expressivity and hypomimia in Parkinson's disease}

Facial motor impairment is a well-recognized manifestation of Parkinson's disease (PD). Hypomimia and facial bradykinesia are commonly reflected in reduced spontaneous movement, diminished expressivity, and slower or less variable facial actions. Because these changes are often judged clinically by observation, several studies have examined whether video-based facial analysis can provide more objective descriptions of PD-related facial behavior. Recent reviews also show that facial expression analysis is increasingly being studied as a candidate digital marker for PD, although available datasets, protocols, and validation settings remain heterogeneous~\cite{oliveira2025facial_expression_review,razzouki2026facial_digital_markers_review}.

Early work established the feasibility of automated facial assessment in PD. Abrami et al.~\cite{abrami2021hypomimia} presented a proof-of-principle study for computer-vision-based hypomimia assessment and discussed its potential relevance for non-invasive monitoring. Bandini et al.~\cite{bandini2017facial} investigated automatic facial-expression analysis in PD, while Jin et al.~\cite{jin2020facial_expression_pd} analyzed facial-expression videos for PD diagnosis using artificial intelligence methods. Su et al.~\cite{su2021hypomimia} studied hypomimia detection from smile videos, and Pegolo et al.~\cite{pegolo2022hypomimia} proposed a face-tracking-based quantitative index for hypomimia. More recently, Novotn\'y et al.~\cite{novotny2022facialbradykinesia} proposed an automated video-based approach for assessing facial bradykinesia in de-novo PD. These studies support the relevance of facial motion analysis, but they are primarily grounded in clinical or semi-controlled recording conditions rather than unconstrained public video.

Recent work has also examined localized facial actions and their relationship to PD-related symptoms. G\'omez et al.~\cite{gomez2023action_units_pd} explored facial-expression and Action Unit domains for PD detection, while Filali Razzouki et al.~\cite{razzouki2025actionunits} analyzed associations between facial Action Units, hypomimia, and clinical scores in early-stage PD. A related study by Filali Razzouki et al.~\cite{razzouki2025au_derivatives} used derivatives of facial Action Units for early-stage PD detection from facial videos. Serb\'ee et al.~\cite{serbee2025hypomimia} studied facial-expression characteristics related to hypomimia using artificial intelligence methods. Across these studies, PD-related facial impairment is treated as a dynamic phenomenon: it involves changes in movement amplitude, timing, variability, and expressivity over time. This motivates temporal analysis rather than relying only on static facial configuration.

\subsection{Interpretable facial-region and action-unit analysis}

In PD facial analysis, interpretability depends on whether the extracted cues plausibly reflect facial movement rather than incidental properties of the recording. This issue is especially relevant in unconstrained videos, where classifiers may exploit identity, age, pose, lighting, video quality, speaking style, head movement, or other dataset-specific factors. The present study therefore focuses on region-level temporal descriptors and uses region-based analyses to examine where discriminative motion information is concentrated.

Facial Action Units are useful in PD facial analysis because they provide localized descriptions of facial muscle-related actions. This interpretation is rooted in the Facial Action Coding System, which decomposes visible facial movement into Action Units~\cite{ekman1978facs}. Modern facial behavior analysis tools such as OpenFace have made automatic extraction of facial landmarks, head pose, gaze, and Action Units more accessible for research applications~\cite{baltrusaitis2018openface}. Prior AU-based PD studies have linked localized facial patterns to hypomimia and clinical measures~\cite{gomez2023action_units_pd,razzouki2025actionunits,razzouki2025au_derivatives}. In this study, we pursue a related form of localized interpretation using the keypoint representation available in YouTubePD rather than clinical AU annotations. Each video is described through 14 predefined facial regions, and the analysis compares static geometry, normalized geometry, velocity, and relative velocity as candidate descriptors for PD-related binary classification.

The region-level analysis is kept deliberately neutral. Although YouTubePD provides region-level annotations~\cite{zhou2023youtubepd_supplement}, the processed files used in this work do not provide a verified mapping from each numerical region index to a named anatomical structure. Region-level findings are therefore reported using identifiers such as Region~07 rather than assigning anatomical labels that cannot be confirmed from the available documentation.

\subsection{Public benchmarks and the YouTubePD dataset}

Public datasets for PD video analysis remain limited because clinical recordings often contain identifiable patient information and sensitive health data. This restricts reproducibility and makes direct comparison across studies difficult. More broadly, recent reviews of computer vision in PD and movement disorders emphasize both the promise of markerless video analysis and the need for careful validation against clinically meaningful targets~\cite{dibiase2025ai_video_pd,pecoraro2025computer_vision_movement}. YouTubePD addresses part of the reproducibility problem by providing a public multimodal benchmark for PD analysis~\cite{zhou2023youtubepd}. The dataset includes in-the-wild videos collected from YouTube together with audio, facial landmarks, and expert annotations. It supports several tasks, including facial-expression-based PD classification, multimodal PD classification, and PD progression synthesis.

The same properties that make YouTubePD valuable also constrain its clinical interpretation. The videos are public and unconstrained rather than standardized clinical examinations. Medication ON/OFF state, time since medication, standardized facial task condition, disease duration, depression or mood state, and full clinical rating context are not generally available. Results obtained on YouTubePD should therefore be interpreted as in-the-wild PD-related video classification or facial-cue analysis, not as medication-controlled severity estimation.

The YouTubePD supplementary material describes region-level annotations and labels for 14 facial regions considered informative for PD analysis~\cite{zhou2023youtubepd_supplement}. This structure motivates the present study. While the benchmark includes visual, audio, and landmark modalities, we focus specifically on temporal descriptors derived from the provided facial-region keypoints. The aim is not to replace the original multimodal benchmark, but to examine what can be learned from its region-level facial representation alone.

\subsection{Static appearance, temporal motion, and sequence modeling}

Static facial representations can capture facial configuration, appearance, and region geometry, but they do not directly describe how facial regions move. This distinction is important for PD, where hypomimia and facial bradykinesia are expressed through reduced or altered movement over time~\cite{bandini2017facial,novotny2022facialbradykinesia,pegolo2022hypomimia}. A single frame may reflect identity, pose, or recording conditions, whereas temporal descriptors can capture changes in region position, size, and shape across the video.

Velocity-based features provide a compact way to represent these changes. Normalization can reduce sensitivity to crop scale, face size, and image framing, while relative velocity can reduce motion components shared across all regions. These representations are particularly relevant for in-the-wild data, where facial motion is mixed with head movement, tracking variation, and recording variability.

Sequence models such as long short-term memory networks and gated recurrent units are commonly used for temporal data~\cite{hochreiter1997lstm,cho2014gru}. However, their advantage depends on the amount and quality of training data. In small and imbalanced benchmarks, compact temporal summaries may offer a more stable alternative to learned sequence models. This study evaluates that trade-off directly by comparing summary-based temporal descriptors with a GRU baseline under the same YouTubePD binary classification protocol.

\subsection{Position of the present study}

This work is an interpretable representation analysis of YouTubePD facial-region dynamics. It does not aim to replace clinical hypomimia assessment, estimate MDS-UPDRS facial-expression scores, or evaluate medication-controlled PD severity. The MDS-UPDRS is a standardized clinical rating scale for PD assessment, and its use requires a clinical context that is not available in YouTubePD~\cite{goetz2008mds_updrs}. Instead, this study asks which temporal facial-region descriptors are most useful for in-the-wild PD-related video classification on a public benchmark.

The study differs from clinical facial bradykinesia and AU-based work in both setting and representation. Clinical studies provide stronger clinical context, but they often rely on restricted datasets and controlled tasks~\cite{abrami2021hypomimia,novotny2022facialbradykinesia,razzouki2025actionunits,razzouki2025au_derivatives}. YouTubePD provides public reproducibility, but lacks several clinical variables~\cite{zhou2023youtubepd}. We therefore use YouTubePD for benchmark-level representation analysis. By comparing raw, normalized, static, velocity, relative-velocity, and GRU-based representations, together with robustness and region-level analyses, the study examines how temporal facial-region dynamics contribute to YouTubePD classification.

\FloatBarrier

%% file: sections/03_materials_methods.tex
\section{Materials and Methods}
\label{sec:materials_methods}

\subsection{Dataset and task formulation}
\label{subsec:dataset}

This study uses the YouTubePD benchmark~\cite{zhou2023youtubepd}, a public multimodal dataset for Parkinson's disease (PD) analysis from in-the-wild videos. The benchmark includes cropped facial videos, audio recordings, facial landmarks, region-level annotations, and expert-provided labels. Although the benchmark supports several tasks, the present work focuses on the facial-region keypoint modality and formulates the problem as binary PD-related video classification.

The original YouTubePD annotations include a control label and non-zero PD-related labels. Since the objective of this study is to analyze facial-motion cues rather than estimate clinical severity, the labels were converted into a binary target:
\begin{equation}
    y =
    \begin{cases}
    0, & \text{if the original label is } 0,\\
    1, & \text{if the original label is greater than } 0.
    \end{cases}
\end{equation}
The term PD-positive is used only with respect to the YouTubePD benchmark annotation. It is not used here as a medication-controlled diagnosis or as a formal clinical severity score.

\subsection{Facial-region representation}
\label{subsec:region_geometry}

The processed YouTubePD keypoint files used in this study provide 14 facial-region entries for each video frame~\cite{zhou2023youtubepd,zhou2023youtubepd_supplement}. These entries are polygonal facial regions rather than dense landmark vectors. Each region is defined by a small set of points, and the number of points may differ across regions. The representation was therefore treated as a region-level geometric description.

\begin{figure}[H]
    \centering
    \includegraphics[width=0.82\linewidth]{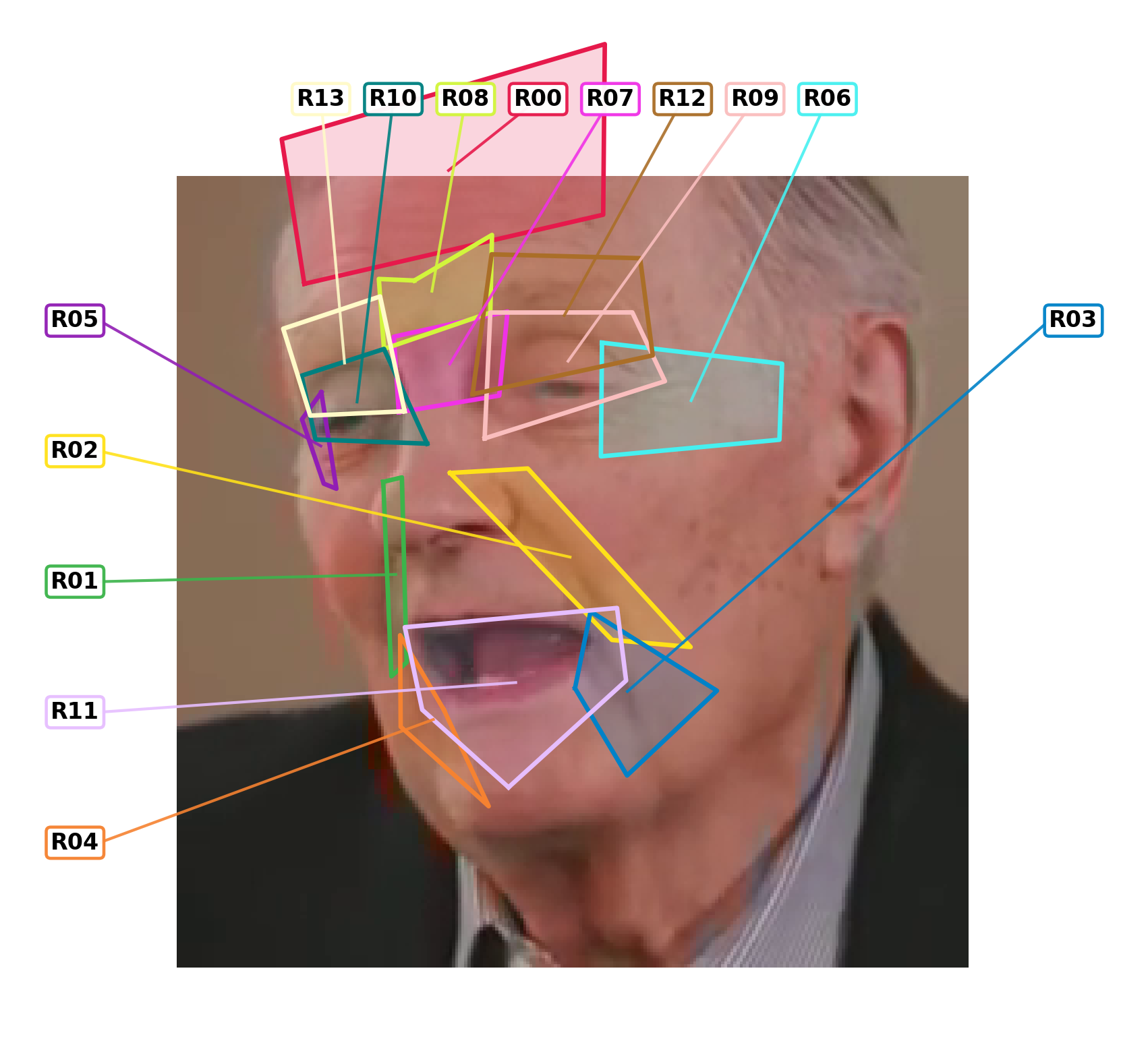}
    \caption{Example visualization of the 14 processed YouTubePD facial-region polygons. The identifiers R00--R13 correspond to the numerical region indices used for feature extraction and region-level analysis.}
    \label{fig:youtubepd_region_layout}
\end{figure}

Figure~\ref{fig:youtubepd_region_layout} shows the numerical region identifiers used in the experiments. A verified anatomical mapping for the 14 processed region indices was not available in the inspected documentation. For this reason, the paper reports region-level findings using numerical identifiers such as R00, R01, and R13 rather than assigning anatomical names.

Let $R_{t,r}$ denote facial region $r$ in frame $t$:
\begin{equation}
    R_{t,r} = \{(x_{t,r,j}, y_{t,r,j})\}_{j=1}^{n_r},
\end{equation}
where $n_r$ is the number of polygon points in region $r$. To obtain a fixed-length representation, each polygon was summarized by descriptors capturing its location, spatial extent, and shape:
\begin{equation}
    \mathbf{g}_{t,r} =
    [
    c_x, c_y,
    x_{\min}, x_{\max},
    y_{\min}, y_{\max},
    w, h, a, \rho
    ].
\end{equation}
Here, $c_x$ and $c_y$ are the region center coordinates, $x_{\min}$ and $x_{\max}$ are the horizontal coordinate bounds, $y_{\min}$ and $y_{\max}$ are the vertical coordinate bounds, $w$ and $h$ are width and height, $a$ is rectangular area, and $\rho$ is aspect ratio.

The center and coordinate bounds were computed as:
\begin{align}
    c_x &= \frac{1}{n_r}\sum_{j=1}^{n_r} x_{t,r,j}, &
    c_y &= \frac{1}{n_r}\sum_{j=1}^{n_r} y_{t,r,j},\\
    x_{\min} &= \min_j x_{t,r,j}, &
    x_{\max} &= \max_j x_{t,r,j},\\
    y_{\min} &= \min_j y_{t,r,j}, &
    y_{\max} &= \max_j y_{t,r,j}.
\end{align}
The remaining descriptors were derived as:
\begin{align}
    w &= x_{\max} - x_{\min}, &
    h &= y_{\max} - y_{\min},\\
    a &= w h, &
    \rho &= \frac{w}{h}.
\end{align}

Each video is then represented as a temporal sequence:
\begin{equation}
    \mathbf{X} \in \mathbb{R}^{T \times 14 \times 10},
\end{equation}
where $T$ is the number of frames, 14 is the number of processed facial regions, and 10 is the number of geometric descriptors extracted from each region.

\subsection{Normalization}
\label{subsec:normalization}

The videos in YouTubePD are collected from unconstrained public sources, so raw coordinates are affected by face scale, crop location, resolution, pose, and camera framing. To reduce these sources of variation, the region descriptors were normalized within each frame using the global spatial extent of all facial regions. For frame $t$, the facial extent was defined as:
\begin{equation}
    W_t = x_{\max}^{(t)} - x_{\min}^{(t)}, \qquad
    H_t = y_{\max}^{(t)} - y_{\min}^{(t)}.
\end{equation}
Horizontal coordinates and widths were divided by $W_t$, vertical coordinates and heights by $H_t$, and area descriptors by $W_tH_t$. Aspect ratio was retained without further scaling. The normalized sequence is denoted by $\mathbf{Z}_{t,r}$.

\subsection{Temporal motion descriptors}
\label{subsec:velocity_descriptors}

To represent facial-region dynamics, first-order velocity descriptors were computed from the normalized sequence:
\begin{equation}
    \Delta \mathbf{Z}_{t,r} =
    \mathbf{Z}_{t,r} - \mathbf{Z}_{t-1,r},
    \qquad t = 2,\ldots,T.
\end{equation}
This captures frame-to-frame changes in region position, size, area, and aspect ratio.

A relative-velocity representation was also derived to reduce movement shared across all regions. For each frame, the average regional velocity was computed as:
\begin{equation}
    \overline{\Delta \mathbf{Z}}_t =
    \frac{1}{14}\sum_{r=1}^{14} \Delta \mathbf{Z}_{t,r}.
\end{equation}
The relative velocity was then defined as:
\begin{equation}
    \Delta \mathbf{Z}^{rel}_{t,r}
    =
    \Delta \mathbf{Z}_{t,r}
    -
    \overline{\Delta \mathbf{Z}}_t.
\end{equation}
This representation was used to examine how much of the velocity-based signal remains after reducing common head, crop, or global facial movement.

\subsection{Video-level temporal summaries}
\label{subsec:temporal_summary_features}

Because the videos have variable lengths, the region-level sequences were converted into fixed-length video descriptors using temporal summaries. Static geometric descriptors were summarized using the mean, standard deviation, minimum, and maximum across frames. Velocity-based descriptors were summarized using the mean, standard deviation, and maximum across frames.

The normalized velocity-only representation therefore contains:
\begin{equation}
    14 \times 10 \times 3 = 420
\end{equation}
features per video. Additional representations were constructed using the same principle to compare raw static-plus-velocity descriptors, normalized static-only descriptors, normalized static-plus-velocity descriptors, and normalized relative-velocity descriptors. The experimental comparisons using these representations are described in Section~\ref{sec:experimental_setup}.

%% file: sections/04_experimental_setup.tex
\section{Experimental Setup}
\label{sec:experimental_setup}

\subsection{Task and data split}
\label{subsec:task_definition}

All experiments were conducted on the binary YouTubePD video classification task~\cite{zhou2023youtubepd}. Videos with label 0 were treated as control samples, while videos with non-zero labels were treated as PD-positive samples according to the benchmark annotation. The official train/test split available with the local YouTubePD files was preserved.

After quality filtering, the final experimental set contained 237 videos. The training split included 71 videos, with 35 control and 36 PD-positive samples. The test split included 166 videos, with 145 control and 21 PD-positive samples. This distribution was considered when selecting the primary evaluation metric and when interpreting class-level results.

\subsection{Quality filtering}
\label{subsec:data_filtering_validation}

Two exclusion criteria were applied before training. Videos with fewer than 30 frames were removed because they provide insufficient temporal information for motion-based descriptors. Samples containing missing feature values were also excluded. These filters removed seven samples in total: one very short video and six videos with missing values. All reported experiments were conducted on the same filtered split.

\subsection{Compared representations}
\label{subsec:compared_representations}

The experiments compare five feature representations derived from the region-level descriptors. Raw static-plus-velocity descriptors were included to assess the effect of using unnormalized coordinates. Normalized static-only descriptors isolated the contribution of facial-region configuration. Normalized static-plus-velocity descriptors combined configuration and motion. Normalized velocity-only descriptors tested whether temporal movement alone was sufficient. Normalized relative-velocity descriptors examined whether region-specific motion remained informative after subtracting movement shared across all regions.

This design directly evaluates the role of normalization, the relative contribution of static and dynamic information, and the degree to which the velocity signal depends on common movement. A GRU sequence baseline was also included to compare fixed-length temporal summaries against direct recurrent sequence modeling~\cite{cho2014gru}.

\subsection{Baseline models}
\label{subsec:classical_baselines}

The fixed-length summary descriptors were evaluated using Logistic Regression, SVM with an RBF kernel, and Random Forest. These models were selected to provide linear, kernel-based, and ensemble baselines under the same feature protocol~\cite{cortes1995support,breiman2001randomforests}. Each model was trained on the training split and evaluated on the held-out test split. The test data were not used during model fitting or model selection.

The best-performing configuration was selected primarily by balanced accuracy. F1 score and AUROC were used as supporting measures because of the imbalanced test distribution~\cite{fawcett2006roc,saito2015precisionrecall,powers2011evaluation}.

\subsection{GRU sequence baseline}
\label{subsec:experimental_gru}

The GRU baseline was trained on normalized velocity sequences. At each timestep, the selected descriptors from all 14 regions were concatenated into a frame-level vector. The model used a bidirectional GRU followed by dropout and a linear classification layer. A validation subset of the training split was used during optimization. This baseline was included to test whether learned temporal sequence modeling improves over summary-based representations under the same YouTubePD split.

\subsection{Relative-motion analysis}
\label{subsec:relative_motion_sanity_check}

The relative-motion experiment was used to help interpret the velocity-based results. In unconstrained videos, facial-region motion may include local facial movement as well as head motion, crop variation, camera movement, and tracking instability. Subtracting the frame-level mean velocity across all regions reduces the common motion component. The relative-velocity representation therefore provides a controlled comparison against the absolute normalized velocity representation.

\subsection{Robustness analysis}
\label{subsec:experimental_seed_robustness}

To examine sensitivity to Random Forest stochasticity, the strongest all-region normalized velocity configuration was repeated across 10 random seeds. The train/test split was fixed in every run, so the reported variation reflects model-level randomness rather than split-level variability. Results are reported as mean and standard deviation across seeds.

\subsection{Region-level ablation}
\label{subsec:experimental_region_ablation}

Region-level ablation was conducted using the normalized velocity representation with Random Forest. The model was first evaluated using all 14 regions and then using each region individually. The all-region setting contains:
\begin{equation}
    14 \times 10 \times 3 = 420
\end{equation}
features per video. Each single-region experiment contains:
\begin{equation}
    1 \times 10 \times 3 = 30
\end{equation}
features per video. Region-level results are reported using the numerical YouTubePD region identifiers because an official anatomical mapping for the processed region indices was not confirmed.

\subsection{Permutation-based interpretability analysis}
\label{subsec:permutation_importance}

To complement the single-region ablation, grouped permutation importance was computed for the all-region normalized velocity Random Forest baseline. The analysis follows the general model-reliance view of variable importance, where the contribution of a feature group is assessed by measuring the performance change after disrupting that information while keeping the trained model fixed~\cite{fisher2019model_reliance}. The analysis was performed on the held-out test split using the same normalized velocity representation described in Section~\ref{subsec:temporal_summary_features}. For each feature group, the corresponding columns were randomly permuted across test videos while the remaining features were kept unchanged. Importance was defined as the decrease in balanced accuracy relative to the unpermuted test performance.

Permutation importance was computed at several levels: region, geometric descriptor, temporal statistic, region--descriptor pair, and individual feature. The procedure was repeated 30 times for each group, and the mean decrease in balanced accuracy was used as the importance estimate. This analysis differs from the single-region ablation: ablation evaluates whether a region is sufficient when used alone, whereas permutation importance estimates how much a feature group contributes within the full all-region model.

\subsection{Evaluation protocol}
\label{subsec:experimental_metrics}

Performance was evaluated on the held-out test split after applying the same quality filters to all methods. Balanced accuracy was used as the primary metric because of the imbalanced test set. Accuracy is reported for completeness. Precision, recall, and F1 score are reported for the PD-positive class, and AUROC is included as a threshold-independent ranking measure. A confusion matrix is reported for the best-performing configuration to support class-level interpretation.

For visual reporting, baseline configurations are summarized using ROC curves and an AUROC--F1 trade-off plot. Region-level ablation results are visualized using mean performance across random seeds with standard-deviation intervals. In the figures, ``Norm.'' denotes normalized descriptors, ``RF'' denotes Random Forest, ``LR'' denotes Logistic Regression, and ``SVM (RBF)'' denotes SVM with an RBF kernel.

%% file: sections/05_results.tex
\section{Results}
\label{sec:results}

\subsection{Overall baseline performance}
\label{subsec:overall_baseline_results}

\begin{figure}[t]
    \centering
    \includegraphics[width=0.85\linewidth]{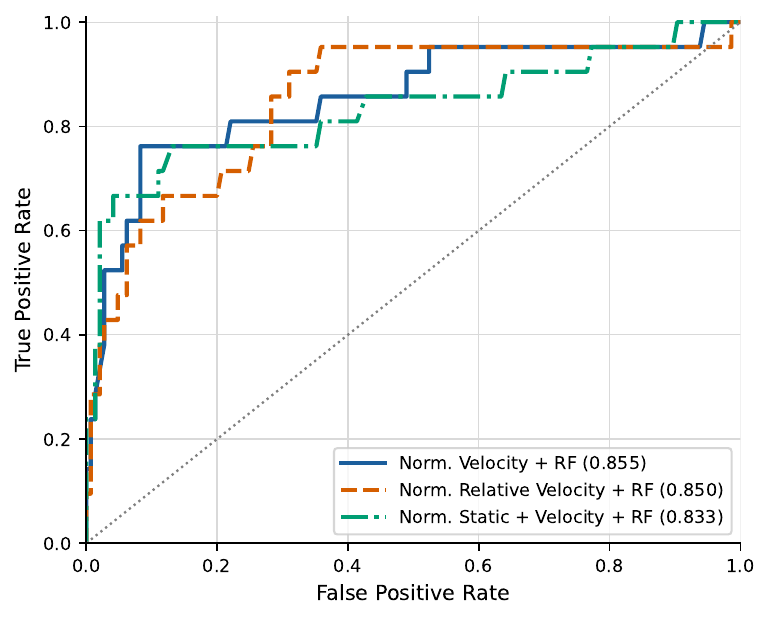}
    \caption{Receiver operating characteristic curves for the strongest baseline configurations on the YouTubePD binary classification task. The curves show sensitivity--specificity trade-offs across decision thresholds and complement the summary metrics reported in Table~\ref{tab:baseline_results}.}
    \label{fig:roc_top_models}
\end{figure}

\begin{figure}[t]
    \centering
    \includegraphics[width=0.72\linewidth]{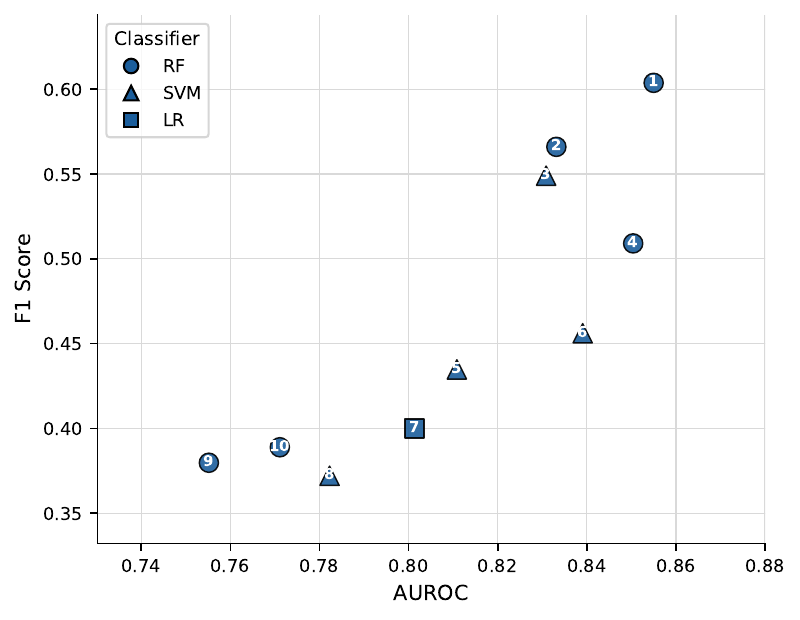}
    \caption{Baseline configurations shown in AUROC--F1 space. Each numbered marker corresponds to the matching configuration order in Table~\ref{tab:baseline_results}. Marker shapes indicate the classifier family.}
    \label{fig:baseline_tradeoff_scatter}
\end{figure}

Table~\ref{tab:baseline_results} summarizes the strongest configurations on the filtered YouTubePD binary classification task. Figure~\ref{fig:roc_top_models} shows the ROC curves for the strongest baseline configurations, while Figure~\ref{fig:baseline_tradeoff_scatter} summarizes the AUROC--F1 trade-off across the compared baselines. The best single-run result was obtained by Random Forest trained on normalized velocity-only descriptors, with accuracy of 0.873, balanced accuracy of 0.826, precision of 0.500, recall of 0.762, F1 score of 0.604, and AUROC of 0.855.

The strongest configuration was obtained using compact motion summaries rather than appearance features or learned sequence representations. This supports the main premise of the study: the processed YouTubePD region keypoints contain useful PD-related information when represented through normalized temporal dynamics. It also shows that the most effective representation in this setting is not the largest descriptor set, but the one most directly focused on region-level motion.

\input{tables/baseline_results_table.tex}

\subsection{Effect of normalization and temporal velocity}
\label{subsec:normalization_velocity_results}

The comparison between raw and normalized descriptors shows that normalization improved performance. The strongest raw static-plus-velocity configuration was below the normalized velocity-only representation across the main metrics. This is consistent with the nature of YouTubePD, where absolute coordinates are affected by face scale, crop position, camera framing, and pose. Normalizing descriptors within each frame reduced this dependence and made the motion representation more robust to recording variability.

The contrast between static and velocity-based descriptors was stronger. Normalized static-only descriptors were clearly weaker than normalized velocity-only descriptors, indicating that frame-level region geometry alone did not capture the most discriminative signal in this benchmark. Velocity descriptors, by contrast, encode how region geometry changes over time, which is more aligned with the facial bradykinesia and hypomimia-related cues motivating this work.

Combining normalized static descriptors with normalized velocity did not improve over velocity alone. The normalized static-plus-velocity Random Forest achieved balanced accuracy of 0.799 and AUROC of 0.833, compared with 0.826 and 0.855 for normalized velocity-only Random Forest. The additional static descriptors may therefore introduce subject- or recording-specific variation that does not improve the binary classification task. In this setting, the more focused temporal representation was preferable to the broader feature set.

\subsection{Relative-velocity analysis}
\label{subsec:relative_velocity_results}

The relative-velocity experiment examined the contribution of common movement across regions. After subtracting the average frame-level velocity, the Random Forest model still achieved balanced accuracy of 0.764, F1 score of 0.509, and AUROC of 0.850. This indicates that useful information remains after reducing shared motion and that the velocity-based result is not explained solely by global movement.

The relative-velocity representation was nevertheless weaker than absolute normalized velocity in balanced accuracy and F1 score. This suggests that the best-performing descriptor captures a mixture of local region-specific dynamics and broader motion patterns present in the videos. In YouTubePD, such broader motion may include head movement, crop behavior, tracking variation, and coordinated facial motion. The velocity-only representation should therefore be interpreted as an in-the-wild facial-motion descriptor rather than a purely local facial muscle measurement.

\subsection{GRU sequence baseline}
\label{subsec:gru_results}

Table~\ref{tab:gru_results} reports the GRU sequence baselines. The GRU trained on normalized velocity sequences achieved balanced accuracy of 0.603, F1 score of 0.282, and AUROC of 0.623. The GRU trained on normalized relative-velocity sequences achieved balanced accuracy of 0.582, F1 score of 0.261, and AUROC of 0.654. Both recurrent baselines were below the summary-based Random Forest configurations.

The result should be interpreted within the limited-data YouTubePD setting rather than as a general conclusion about recurrent sequence models. After validation splitting, the GRU was optimized on a small number of videos, while the Random Forest used compact video-level summaries. Under this protocol, summary descriptors provided a more reliable baseline than direct recurrent sequence modeling.

\input{tables/gru_results_table}

\subsection{Robustness of the strongest representation}
\label{subsec:seed_robustness_results}

The all-region normalized velocity Random Forest was repeated across 10 random seeds. As shown in Table~\ref{tab:seed_robustness}, it achieved balanced accuracy of $0.810 \pm 0.018$, F1 score of $0.592 \pm 0.022$, and AUROC of 0.855 $\pm$ 0.005. The small variation in AUROC indicates that the ranking performance of this representation was not dependent on a single favorable Random Forest initialization.

The threshold-dependent metrics varied more than AUROC, which is expected with only 21 PD-positive videos in the test set. A small change in the number of correctly detected PD-positive samples can noticeably affect recall and F1 score. Even so, the robustness results support the same conclusion as the single-run comparison: normalized velocity descriptors provide the most stable representation among the tested configurations.

\input{tables/seed_robustness_table}

\subsection{Region-level ablation}
\label{subsec:region_ablation_results}

\begin{figure}[t]
    \centering
    \includegraphics[width=0.82\linewidth]{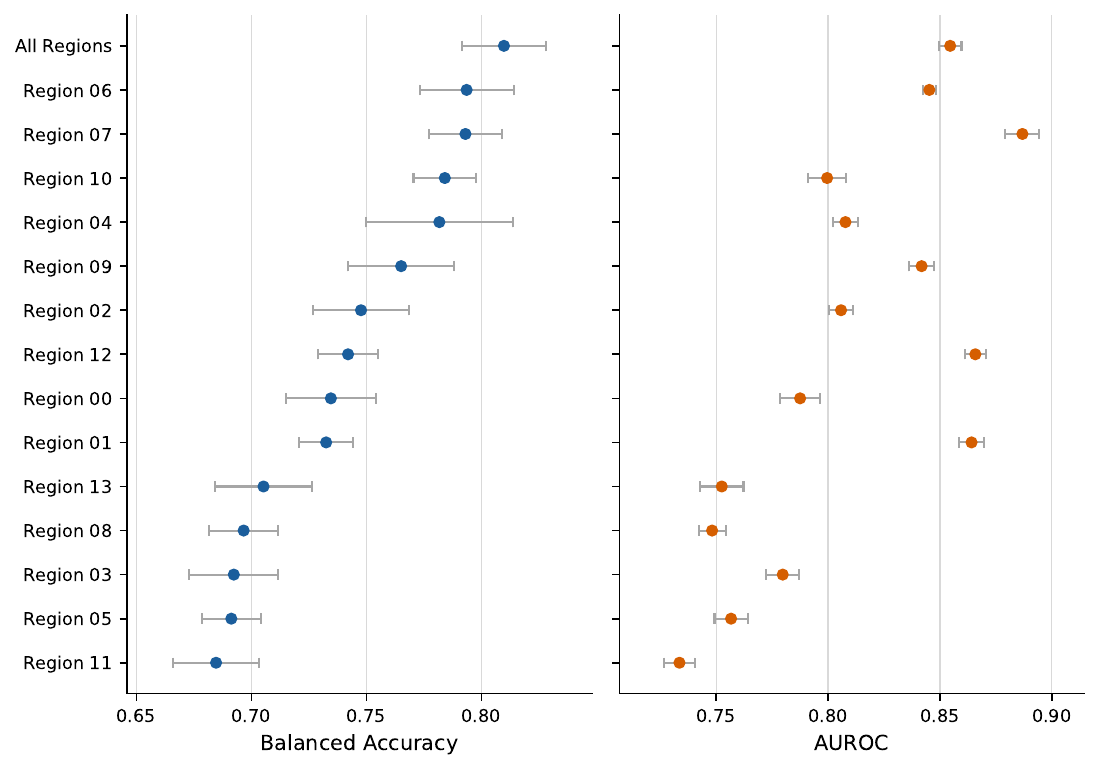}
    \caption{Single-region ablation results on the YouTubePD binary classification task. Each point shows the performance of a model trained using one facial region only, while the full-region configuration is included for reference. Error bars indicate standard deviation across random seeds. The figure summarizes how discriminative performance varies across the predefined facial regions.}
    \label{fig:region_ablation_dotplot}
\end{figure}

Figure~\ref{fig:region_ablation_dotplot} summarizes the region-level ablation, and Table~\ref{tab:region_ablation_top} lists the strongest region groups. The all-region representation achieved the highest mean balanced accuracy across seeds, although several single-region configurations remained competitive despite using only 30 features per video.

Among the individual regions, Region 07 produced the highest mean AUROC, while Region 06 achieved one of the strongest balanced-accuracy results. Regions 12 and 01 also showed relatively strong AUROC values. These results indicate that discriminative information is not distributed uniformly across the processed YouTubePD regions.

The ablation analysis also supports the interpretability of the baseline. The classifier does not appear to rely only on weak contributions aggregated across all regions; instead, some regions retain stronger temporal motion cues when evaluated in isolation. Because the processed region indices could not be mapped reliably to anatomical facial structures, these findings are reported using numerical region identifiers.

\input{tables/region_ablation_top_table}

\subsection{Permutation importance analysis}
\label{subsec:results_permutation_importance}

Permutation importance provided a more detailed interpretation of the normalized velocity Random Forest baseline. The retrained model used for this analysis achieved accuracy of 0.880, balanced accuracy of 0.829, F1 score of 0.615, and AUROC of 0.849, which is consistent with the primary baseline results reported in Table~\ref{tab:baseline_results}. The interpretation is therefore based on a representative instance of the best-performing representation.

At the region level, Region 10 produced the largest decrease in balanced accuracy when permuted ($0.060 \pm 0.023$), followed by Region 07 ($0.042 \pm 0.011$) and Region 06 ($0.029 \pm 0.015$). This partly agrees with the single-region ablation, where Region 07 and Region 06 were among the strongest individual regions. The difference in ranking is expected because the two analyses answer different questions: the ablation tests whether a region is informative in isolation, whereas permutation importance measures the contribution of a region within the full model.

Descriptor-level permutation indicated that area ($0.064 \pm 0.023$), height ($0.050 \pm 0.024$), $x_{\max}$ ($0.039 \pm 0.014$), $y_{\min}$ ($0.034 \pm 0.015$), and width ($0.032 \pm 0.013$) were the most influential descriptor groups. These results suggest that the classifier relied more strongly on changes in regional extent and shape than on centroid motion alone.

\begin{figure}[t]
    \centering
    \includegraphics[width=0.70\linewidth]{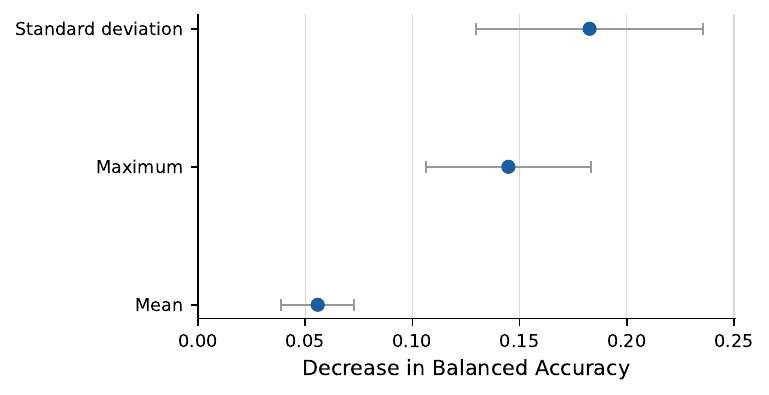}
    \caption{Grouped permutation importance by temporal statistic for the normalized velocity Random Forest baseline. Importance is measured as the decrease in balanced accuracy after permuting each statistic group on the held-out test set. Points indicate mean importance and horizontal intervals indicate standard deviation across repeated permutations.}
    \label{fig:permutation_importance_statistic}
\end{figure}

Figure~\ref{fig:permutation_importance_statistic} summarizes the temporal-statistic importance. The standard deviation of velocity produced the largest decrease in balanced accuracy ($0.183 \pm 0.053$), followed by maximum velocity ($0.145 \pm 0.038$) and mean velocity ($0.056 \pm 0.017$). This indicates that variability and peak changes in facial-region motion contributed more strongly than average motion.

The highest-ranked individual features were concentrated in selected region--descriptor combinations, including Region 10 height variability, Region 07 height variability, Region 06 $y_{\min}$ variability, and Region 07 area variability. These findings support the region-ablation results and show that the discriminative signal is concentrated in specific temporal dynamics rather than being uniformly distributed across all region descriptors.

\subsection{Confusion matrix analysis}
\label{subsec:confusion_matrix_results}

\begin{figure}[t]
    \centering
    \includegraphics[width=0.55\linewidth]{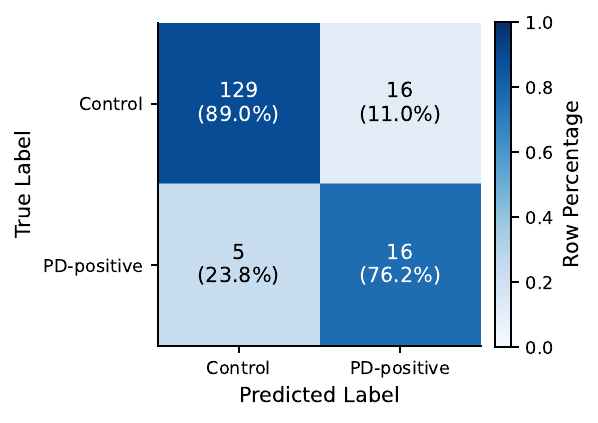}
    \caption{Confusion matrix for the best-performing Normalized Velocity + Random Forest baseline on the held-out YouTubePD test split. Cell values report counts and row percentages.}
    \label{fig:confusion_matrix}
\end{figure}

Figure~\ref{fig:confusion_matrix} shows the confusion matrix for the normalized velocity Random Forest baseline. The model correctly classified 129 of 145 control videos and 16 of 21 PD-positive videos, with 16 false positives and 5 false negatives. This corresponds to recall of 0.762 and precision of 0.500 for the PD-positive class.

The model therefore detected most PD-positive videos in the test set but produced a non-negligible number of false positives. This behavior is consistent with the imbalanced test distribution and the emphasis on balanced accuracy and PD-positive recall. In this study, the confusion matrix is interpreted only within the YouTubePD benchmark setting. It supports the conclusion that normalized facial-region velocity contains useful classification cues, while also showing that precision and specificity remain limited.

\subsection{Summary of findings}
\label{subsec:results_summary}

The results support five main observations. First, normalized velocity descriptors were the strongest representation among the tested configurations. Second, temporal motion was more informative than static facial-region geometry. Third, the GRU sequence baseline did not outperform compact summary descriptors under the available YouTubePD training split. Fourth, region-level ablation showed that discriminative information is unevenly distributed across the processed facial-region indices. Fifth, permutation importance showed that the strongest model relied mainly on velocity variability and peak motion, especially in descriptor groups related to regional extent and shape.

Together, these findings support the central claim of this study: lightweight normalized temporal facial-region motion descriptors provide useful, stable, and interpretable cues for in-the-wild PD-related video classification on YouTubePD.

\FloatBarrier

%% file: tables/baseline_results_table.tex
\begin{table}[H]
\centering
\caption{Top baseline configurations on the YouTubePD binary classification task. Norm. denotes normalized descriptors; RF denotes Random Forest; LR denotes Logistic Regression; SVM (RBF) denotes Support Vector Machine with a radial basis function kernel.}
\label{tab:baseline_results}
\begin{tabular}{clccccc}
\toprule
No. & Configuration & Bal. Acc. & F1 & AUROC & Precision & Recall \\
\midrule
1 & Norm. Velocity + RF & 0.826 & 0.604 & 0.855 & 0.500 & 0.762 \\ 
2 & Norm. Static + Velocity + RF & 0.799 & 0.566 & 0.833 & 0.469 & 0.714 \\ 
3 & Norm. Velocity + SVM (RBF) & 0.778 & 0.549 & 0.831 & 0.467 & 0.667 \\ 
4 & Norm. Relative Velocity + RF & 0.764 & 0.509 & 0.850 & 0.412 & 0.667 \\ 
5 & Norm. Static + Velocity + SVM (RBF) & 0.743 & 0.435 & 0.811 & 0.312 & 0.714 \\ 
6 & Norm. Relative Velocity + SVM (RBF) & 0.730 & 0.456 & 0.839 & 0.361 & 0.619 \\ 
7 & Raw Static + Velocity + LR & 0.723 & 0.400 & 0.801 & 0.278 & 0.714 \\ 
8 & Raw Static + Velocity + SVM (RBF) & 0.712 & 0.372 & 0.782 & 0.246 & 0.762 \\ 
9 & Norm. Static + RF & 0.709 & 0.380 & 0.755 & 0.259 & 0.714 \\ 
10 & Raw Static + Velocity + RF & 0.706 & 0.389 & 0.771 & 0.275 & 0.667 \\ 
\bottomrule
\end{tabular}
\end{table}

%% file: tables/gru_results_table.tex
\begin{table}[t]
\centering
\caption{GRU sequence baselines. The GRU was trained using a validation subset of the official training split and evaluated on the same test set as the tabular baselines.}
\label{tab:gru_results}
\begin{tabular}{lcccc}
\toprule
Configuration & Bal. Acc. & F1 & AUROC & Best Epoch \\
\midrule
GRU Norm. Velocity & 0.603 & 0.282 & 0.623 & 20 \\ 
GRU Norm. Relative Velocity & 0.582 & 0.261 & 0.654 & 29 \\ 
\bottomrule
\end{tabular}
\end{table}

%% file: tables/seed_robustness_table.tex
\begin{table}[t]
\centering
\caption{Seed robustness of the all-region normalized velocity Random Forest baseline over 10 random seeds.}
\label{tab:seed_robustness}
\begin{tabular}{lcc}
\toprule
Metric & Mean & Std. \\
\midrule
Accuracy & 0.874 & 0.007 \\
Balanced Accuracy & 0.810 & 0.018 \\
F1 Score & 0.592 & 0.022 \\
AUROC & 0.855 & 0.005 \\
Precision & 0.502 & 0.019 \\
Recall & 0.724 & 0.038 \\
\bottomrule
\end{tabular}
\end{table}

%% file: tables/region_ablation_top_table.tex
\begin{table}[t]
\centering
\caption{Top region-level ablation results using normalized velocity descriptors and Random Forest over 10 seeds. Region names are kept as YouTubePD indices because the anatomical mapping was not confirmed.}
\label{tab:region_ablation_top}
\begin{tabular}{lccc}
\toprule
Region Group & Bal. Acc. & F1 & AUROC \\
\midrule
All Regions & 0.810 $\pm$ 0.018 & 0.592 $\pm$ 0.022 & 0.855 $\pm$ 0.005 \\ 
Region 06 & 0.794 $\pm$ 0.020 & 0.539 $\pm$ 0.033 & 0.845 $\pm$ 0.003 \\ 
Region 07 & 0.793 $\pm$ 0.016 & 0.619 $\pm$ 0.024 & 0.887 $\pm$ 0.008 \\ 
Region 10 & 0.784 $\pm$ 0.014 & 0.498 $\pm$ 0.020 & 0.800 $\pm$ 0.008 \\ 
Region 04 & 0.782 $\pm$ 0.032 & 0.495 $\pm$ 0.033 & 0.808 $\pm$ 0.006 \\ 
Region 09 & 0.765 $\pm$ 0.023 & 0.519 $\pm$ 0.030 & 0.842 $\pm$ 0.006 \\ 
Region 02 & 0.748 $\pm$ 0.021 & 0.486 $\pm$ 0.024 & 0.806 $\pm$ 0.005 \\ 
Region 12 & 0.742 $\pm$ 0.013 & 0.496 $\pm$ 0.022 & 0.866 $\pm$ 0.005 \\ 
Region 00 & 0.735 $\pm$ 0.020 & 0.455 $\pm$ 0.024 & 0.788 $\pm$ 0.009 \\ 
Region 01 & 0.733 $\pm$ 0.012 & 0.485 $\pm$ 0.021 & 0.864 $\pm$ 0.005 \\ 
\bottomrule
\end{tabular}
\end{table}

%% file: sections/06_discussion.tex
\section{Discussion}
\label{sec:discussion}

This study investigated whether temporal facial-region dynamics extracted from YouTubePD keypoints provide useful information for in-the-wild PD-related video classification. The main finding is that normalized velocity descriptors produced the strongest and most stable performance among the evaluated representations. This suggests that, within the YouTubePD benchmark, facial-region movement over time is more informative than static facial-region geometry alone.

\subsection{Importance of temporal facial-region motion}

The comparison between static and velocity-based representations indicates that the discriminative signal in YouTubePD is primarily dynamic. Static geometry describes the average position, size, and shape of facial regions, but it does not directly represent how those regions move. In contrast, velocity features capture frame-to-frame changes in region location and geometry. This is more consistent with the nature of hypomimia and facial bradykinesia, which are expressed through reduced or altered movement rather than through a single static facial configuration~\cite{novotny2022facialbradykinesia,abrami2021hypomimia,bandini2017facial}.

The strongest performance was achieved by normalized velocity-only features, rather than by
the combination of static and velocity features. This suggests that adding static geometry does not necessarily improve classification and may introduce subject- or recording-specific variation.

\subsection{Role of normalization}

Normalization also played a central role in improving performance. Raw coordinate-based features are affected by crop size, face scale, image resolution, and head position. These factors are especially problematic in YouTubePD because the videos are collected from public sources and are not recorded under a standardized protocol~\cite{zhou2023youtubepd}. By normalizing region geometry within each frame, the representation becomes less dependent on absolute coordinate values and more focused on relative facial-region behavior. The improvement from raw features to normalized velocity features suggests that scale and position variation can obscure useful PD-related motion cues. Future work on in-the-wild PD video analysis should therefore avoid relying directly on unnormalized coordinates when region-level motion descriptors are available. Even simple normalization can substantially improve the usefulness of facial-region motion descriptors.

\subsection{Global motion and relative velocity}

The relative-velocity experiment was included to test whether the model was relying only on global movement, such as head motion, camera motion, or crop instability. The performance of relative velocity was lower than that of absolute normalized velocity, but it remained meaningful. This result suggests that the strongest velocity representation contains both local facial-region dynamics and broader motion patterns.

This finding should be interpreted carefully. On one hand, the relative-velocity result shows that region-specific motion still carries discriminative information after subtracting common movement across regions. On the other hand, the performance drop indicates that shared motion contributes to the best result. In in-the-wild videos, such shared motion may include clinically relevant movement patterns, but it may also include non-clinical factors such as head movement or tracking variation. Therefore, the current representation should be viewed as a useful in-the-wild facial-motion descriptor rather than as a pure measure of local facial muscle movement.

\subsection{GRU sequence modeling under limited training data}

The GRU sequence baseline did not outperform the summary-based classical models. This result should be interpreted in light of the limited YouTubePD training split. With only 71 training videos after filtering, the GRU baseline had limited data from which to learn stable temporal patterns, particularly after reserving part of the training split for validation.

The stronger performance of Random Forest on compact temporal summaries suggests that, for small in-the-wild facial datasets, carefully designed summary descriptors may be more reliable than learned sequence models. This does not imply that recurrent or transformer-based models are unsuitable for PD facial analysis in general. Rather, it indicates that, under the present YouTubePD setting, model complexity must be balanced against data availability. A simple model with meaningful motion descriptors can outperform a more flexible temporal model when training data are limited.

\subsection{Region-level interpretation}
\label{subsec:discussion_region_interpretation}

The region-level ablation and permutation-importance analyses provide complementary views of the learned representation. The ablation experiment evaluates whether an individual region is sufficient for classification when used alone. In contrast, permutation importance estimates how much a region or descriptor group contributes within the full all-region model. The two analyses are therefore not expected to produce identical rankings.

In the ablation analysis, the all-region representation achieved the strongest mean balanced accuracy, while several individual regions remained competitive despite using only 30 features per video. Region 07 and Region 06 were among the strongest single-region settings. In the full-model permutation analysis, Region 10 produced the largest importance value, followed by Region 07 and Region 06. This suggests that some regions may be informative individually, whereas others contribute more through their interaction with the remaining facial-region dynamics.

The descriptor- and statistic-level permutation results further refine this interpretation. The most influential descriptor groups were related to regional extent and shape, including area, height, width, and coordinate bounds. Among temporal summaries, velocity standard deviation was more influential than maximum or mean velocity. This suggests that variability in facial-region motion is an important component of the discriminative signal in YouTubePD.

These findings should remain computational rather than anatomical. The processed files used in this study did not provide a verified mapping from numerical region indices to named facial structures. For this reason, the results are reported using neutral region identifiers. A reliable anatomical mapping would allow future work to connect these region-level patterns more directly to clinical interpretations of facial bradykinesia and hypomimia.

\subsection{Implications for YouTubePD-based research}

The results suggest that YouTubePD can support more than appearance-based or multimodal classification. Its region-level keypoint representation also enables interpretable temporal motion analysis~\cite{zhou2023youtubepd,zhou2023youtubepd_supplement}. The proposed pipeline shows that simple normalized velocity descriptors can capture meaningful PD-related cues while remaining lightweight and reproducible.

This is useful for future YouTubePD studies for two reasons. First, it provides a strong non-deep baseline for the facial-region keypoint modality. Any more complex temporal model should be compared against normalized velocity summaries, not only against static or raw-coordinate features. Second, it highlights the importance of representation choice. The difference between static, raw, normalized, velocity, and relative-velocity features was substantial, indicating that preprocessing and feature design can influence results as much as model choice.

\subsection{Clinical interpretation}

Although the findings are promising for in-the-wild video classification, they should not be interpreted as evidence of medication-controlled clinical PD assessment. YouTubePD does not generally provide medication ON/OFF state, time since medication, standardized facial task conditions, disease duration, depression or mood state, or full clinical rating context. These factors are important for clinical interpretation of facial expressivity and hypomimia~\cite{novotny2022facialbradykinesia,razzouki2025actionunits}.

Therefore, the proposed method should be viewed as an interpretable benchmark approach for detecting PD-related facial cues in YouTubePD videos. It does not estimate clinical hypomimia severity, replace neurological examination, or provide medication-independent assessment. A clinical validation study would require standardized recordings, subject-level clinical metadata, and medication-state information. The value of the present work lies in showing which facial-region dynamics are informative in a public in-the-wild benchmark and in providing a reproducible baseline for future studies.

\subsection{Summary}
\label{subsec:discussion_summary}

Overall, the discussion supports the central conclusion that normalized temporal motion is a more effective representation than static geometry for YouTubePD facial-region classification. The findings also show that classical models trained on compact motion summaries can outperform GRU sequence modeling in this limited-data setting. Region-level ablation and permutation importance provide additional evidence that discriminative information is concentrated unevenly across the predefined YouTubePD regions and is driven mainly by temporal variability and peak changes in region geometry. Together, these results position normalized facial-region velocity as a useful, interpretable, and reproducible representation for in-the-wild PD-related video classification.

%% file: sections/07_limitations.tex
\section{Limitations}
\label{sec:limitations}

This study has several limitations that should be considered when interpreting the results. The first and most important limitation is related to the nature of the YouTubePD dataset. YouTubePD provides a valuable public benchmark for in-the-wild PD video analysis, but it is not a controlled clinical assessment dataset~\cite{zhou2023youtubepd}. The videos are collected from public YouTube sources and therefore vary in recording conditions, lighting, pose, facial visibility, speaking behavior, camera quality, and video context. These factors may influence the extracted facial-region dynamics and may contribute to the classification signal.

A second limitation is the lack of detailed clinical metadata. In particular, medication ON/OFF state, time since medication, disease duration, standardized facial task condition, depression or mood state, and detailed clinical rating context are not generally available. These variables are important for clinical interpretation of facial expressivity and hypomimia~\cite{novotny2022facialbradykinesia,razzouki2025actionunits}. For this reason, the results should not be interpreted as medication-controlled clinical severity assessment or as estimation of MDS-UPDRS facial-expression scores~\cite{goetz2008mds_updrs}. The findings are limited to PD-related video classification and facial cue analysis within the YouTubePD benchmark.

Another limitation concerns the dataset size and class distribution. After filtering, the binary task contained 237 videos, with 71 training samples and 166 test samples. Although the training set was approximately balanced, the test set was highly imbalanced, containing 145 control videos and 21 PD-positive videos. This imbalance makes accuracy alone insufficient and increases the uncertainty of metrics that depend heavily on the small PD-positive test set~\cite{saito2015precisionrecall,powers2011evaluation}. For this reason, balanced accuracy, F1 score, recall, precision, AUROC, and confusion matrices were reported. Nevertheless, the limited number of PD-positive test samples should be considered when interpreting performance.

The study also relies on the region-level keypoint representation provided with YouTubePD. Each frame is represented by 14 predefined facial regions rather than by dense facial landmarks or directly measured facial muscle activations. This representation is useful for lightweight and interpretable analysis, but it limits the level of anatomical detail that can be extracted. In addition, the processed files used in this study did not provide a fully verified mapping between the numerical region indices and named anatomical facial structures. Consequently, the region-level ablation results are reported using neutral identifiers such as Region~00 and Region~07. These results indicate that discriminative information is unevenly distributed across YouTubePD regions, but they should not be interpreted as direct anatomical or clinical findings.

A further limitation is that the proposed descriptors may still capture non-facial sources of motion. Although relative velocity was evaluated to reduce common global movement, normalized velocity features may still contain information related to head movement, face-crop variation, camera motion, or landmark-tracking instability. In unconstrained videos, these factors are difficult to separate completely from local facial movement. The relative-velocity experiment suggests that region-specific motion remains informative, but it also indicates that shared movement contributes to the strongest representation.

Finally, the GRU baseline was evaluated under the same limited-data setting as the classical baselines. Its weak performance should therefore be interpreted cautiously. The result does not imply that sequence models are unsuitable for PD facial analysis in general. Rather, it suggests that, for the available YouTubePD binary split, compact temporal summaries are more reliable than a recurrent model trained on a small number of videos. Larger datasets, stronger temporal supervision, or pretraining strategies may change this conclusion.

Overall, these limitations define the scope of the present study. The proposed framework should be viewed as a reproducible and interpretable analysis of temporal facial-region dynamics for YouTubePD-based in-the-wild PD video classification. Stronger clinical claims would require controlled recordings, richer clinical metadata, medication-state information, and external validation on clinical datasets.

%% file: sections/08_conclusion.tex
\section{Conclusion}
\label{sec:conclusion}

This paper presented an interpretable temporal facial-region dynamics framework for YouTubePD binary PD-related video classification. Using the region-level keypoints provided with the benchmark, each video was represented through geometric descriptors extracted from 14 predefined facial regions. The study compared raw, normalized, static, velocity-based, relative-velocity, and GRU-based representations under a consistent experimental protocol.

The main finding is that normalized velocity descriptors provide the strongest and most stable representation among the tested configurations. Random Forest trained on normalized velocity-only summaries outperformed static geometry, raw descriptors, relative-velocity variants, and GRU sequence baselines. Robustness analysis confirmed that the result was stable across Random Forest seeds, while region-level ablation and permutation importance showed that discriminative information is unevenly distributed across the YouTubePD facial regions. The permutation analysis further indicated that the strongest full-model contributions were associated with temporal variability and peak changes in region geometry, particularly descriptors related to regional extent and shape.

These results support the use of lightweight temporal facial-region motion descriptors for in-the-wild PD-related video classification. The framework provides a reproducible representation-level baseline for YouTubePD and offers a more interpretable alternative to purely appearance-based or high-capacity temporal models in this limited-data setting. At the same time, the findings should remain within the benchmark scope. The study does not claim clinical diagnosis, medication-state estimation, MDS-UPDRS facial-expression scoring~\cite{goetz2008mds_updrs}, or medication-controlled severity assessment.

Future work should validate the representation on clinical datasets with standardized facial tasks, subject-level metadata, and medication-state information. A verified mapping between YouTubePD region indices and anatomical facial regions would also allow stronger clinical interpretation of the region-level and permutation-importance findings. In addition, future studies could examine whether the normalized velocity representation remains useful when combined with stronger temporal models, multimodal cues, or external clinical validation cohorts.

%% file: sections/data_code_availability.tex
\section*{Data and Code Availability}

The preprocessing and experiment code, including configuration files and scripts required to reproduce dataset preparation, task construction, baseline training, robustness analysis, and figure generation, will be made available in a public repository upon publication.

%% file: sections/conflict_of_interest.tex
\section*{Conflict of Interest}

The author declares no conflict of interest.

%% file: sections/acknowledgments.tex
\section*{Acknowledgments}

The author thanks the creators of the YouTubePD benchmark for making an in-the-wild PD analysis dataset publicly available.